\documentclass[10pt,twocolumn,letterpaper]{article}

\usepackage{cvpr}
\usepackage{times}
\usepackage{epsfig}
\usepackage{graphicx}
\usepackage{amsmath}
\usepackage{amssymb}
\usepackage{multirow}
\usepackage{subfig}

\usepackage{tabularx,colortbl}

\usepackage{booktabs}
\usepackage[linesnumbered,noend,ruled,oldcommands]{algorithm2e}

\usepackage[pagebackref=true,breaklinks=true,letterpaper=true,colorlinks,bookmarks=false]{hyperref}

\newcommand{\x}{\mathbf{x}}
\newcommand{\p}{\mathbf{p}}

\makeatletter

\DeclareRobustCommand\onedot{\futurelet\@let@token\@onedot}
\def\@onedot{\ifx\@let@token.\else.\null\fi\xspace}
 
\def\ie{\emph{i.e}\onedot} 
 
\def\etc{\emph{etc}\onedot} 
 
\def\etal{\emph{et al}\onedot}

\makeatother
 \cvprfinalcopy 

\ifcvprfinal\fi
\begin{document}

\title{Object-centric Sampling for Fine-grained Image Classification}

\author{
	Xiaoyu Wang $^{*}$\,\,\, \,\,\,
	Tianbao Yang$^{*}\dagger$\,\,\,\,\,\,
	Guobin Chen$^{*}\ddagger$\,\,\,\,\,\,
	Yuanqing Lin$^{*}$\,\,\,\,\,\,\\\\
$^*$NEC Labs America\,\,\,\,\,\,\,\,\,\,\,\, $\dagger$ University of Iowa \,\,\,\,\,\,\,\,\,\,\,$\ddagger$University of Missouri\\\\
{\tt\small fanghuaxue@gmail.com}\,\,\,
{\tt\small tianbao-yang@uiowa.edu}\,\,\,
{\tt\small gcn38@mail.missouri.edu}\,\,\,
{\tt\small ylin@nec-labs.com}\,\,\,
}
\maketitle
\begin{abstract}

This paper proposes to go beyond the state-of-the-art deep convolutional neural network (CNN) by
incorporating the information from object detection, focusing on dealing with fine-grained image
classification. Unfortunately, CNN suffers from  over-fiting when it is trained on existing
fine-grained image classification benchmarks, which typically only consist of less than a few tens
of thousands training images. Therefore, we first construct a large-scale fine-grained car
recognition dataset that consists of 333 car classes with more than 150 thousand training images.
With this large-scale dataset, we are able to build a strong baseline for CNN with top-1
classification accuracy of 81.6\%. One major challenge in fine-grained image classification is that
many classes are very similar to each other while having large within-class variation. One
contributing factor to the within-class variation is cluttered image background.  However, the
existing CNN training takes uniform window sampling over the image, acting as blind on the location
of the object of interest. In contrast, this paper proposes an \emph{object-centric sampling} (OCS)
scheme that samples image windows based on the object location information. The challenge in using
the location information lies in how to design powerful object detector and how to handle the
imperfectness of detection results. To that end, we design a saliency-aware object detection
approach specific for the setting of fine-grained image classification, and the uncertainty of
detection results are naturally handled in our OCS scheme. Our framework is demonstrated to be very
effective, improving top-1 accuracy to 89.3\% (from 81.6\%) on the large-scale fine-grained car
classification dataset. 

\end{abstract}

\section{Introduction}

\begin{figure}[tr]
\includegraphics[width=1.0\linewidth]{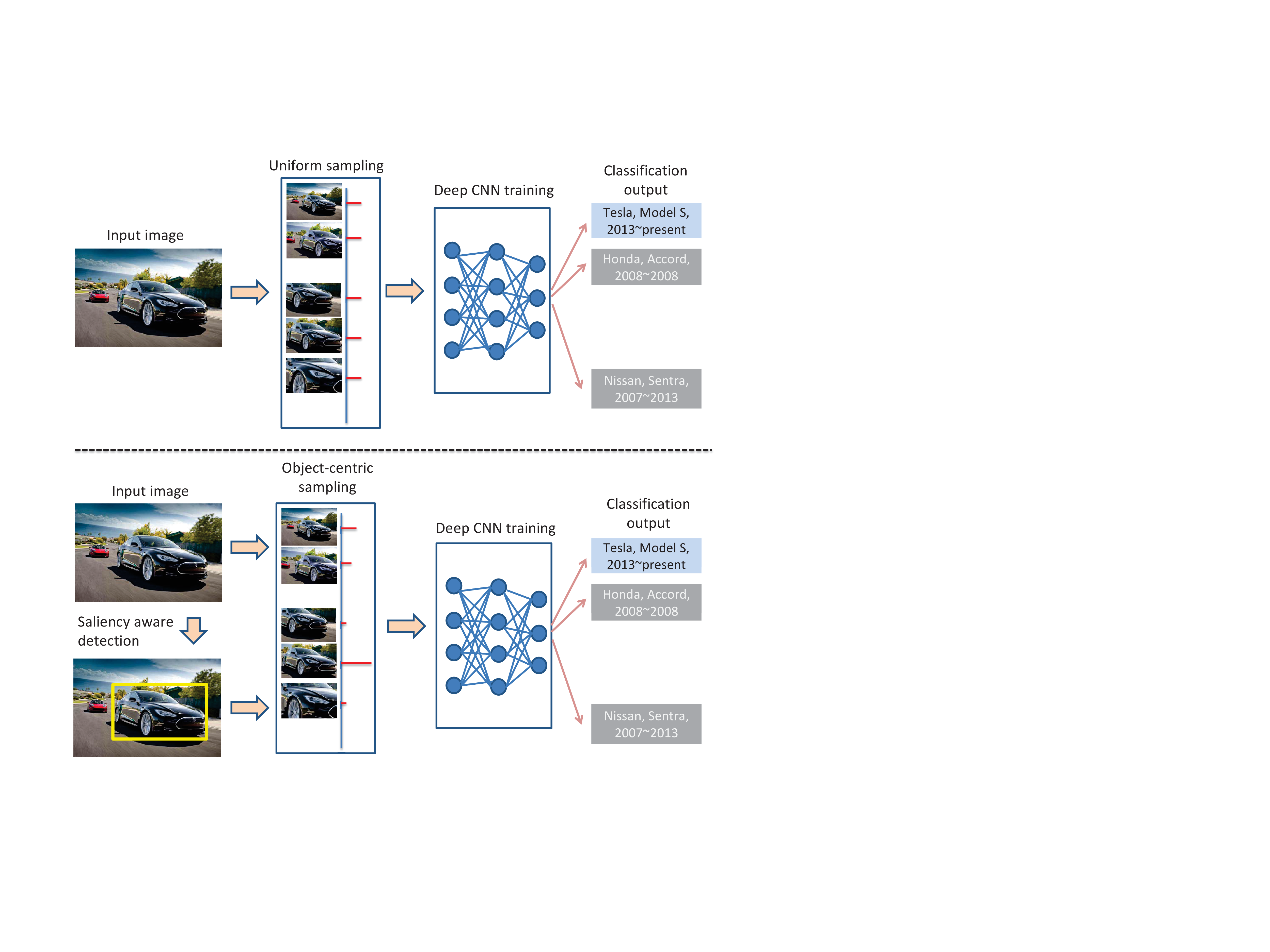} 

\caption{A conventional CNN pipeline with uniform sampling (upper figure) vs our proposed CNN
pipeline with object-centric sampling (lower figure). The red bars in the second column in both
figures indicate the importance of the sampled image windows. The key for object-centric sampling
scheme to help classification is in two folds: 1) how to obtains accurate detection results; and 2)
how to deal with imperfectness of detection results. For the former, we propose a saliency-aware
detection based on Regionlet detection framework; for the later, we propose a multinomial sampling
scheme that softly emphasizes both detection results and original images in windows sampling.}

\label{fig:pull_fig}
\end{figure}

Large-scale image classification has been undergoing amazing progress since the seminal work of
Krizhevsky \etal~\cite{Alex2012} in 2012, which trained deep convolutional neural networks (CNN) to
produce dramatic classification accuracy on the ImageNet Large Scale Visual Recognition Challenge
(ILSVRC2012). In contrast, the success of deep CNN on fine-grained image classification has not been
so overwhelming. One reason for this slackness is that CNN often requires large-scale training data
to avoid over-fitting but fine-grained image labels are expensive to obtain. For example, Amazon
Mechanical Turkers may be easy to tell whether an image contains a car, but it would be very
difficult for them to tell whether an image contains a Chevrolet Equinox 2010$\sim$2013. In fact,
most existing fine-grained image classification benchmark datasets only consist of a few tens of
thousands (or less) of training images. For example, the Stanford car dataset~\cite{KrauseICCV13}
only consists of 8,144 training images. As a result, training CNN on this dataset only obtains 68\%
top-1 classification accuracy (compared to 69.5\% top-1 accuracy using conventional LLC features
with SPM and SVM\cite{KrauseICCV13}). 

To build a strong baseline for CNN so that we can study any further improvement on CNN, we construct
a large-scale fine-grained car classification dataset. Our car dataset consists of 157,023 training
images and 7,840 testing images from 333 categories, each of which is described as "which brand,
which model and which year blanket". A year blanket means the year range that a car model does not
change its outlook design (so that the cars in the same year blanket are not visually
distinguishable). The training set of this dataset is more than one magnitude larger than the
Stanford car dataset~\cite{KrauseICCV13}. This enables us to build a strong baseline performance
using CNN, achieving 81.6\% top-1 classification accuracy (compared to 52.8\% top-1 accuracy using
conventional LLC features with SPM and SVM according to our own experiment).

One major challenge in fine-grained image classification is to distinguish subtle between-class
differences while each class often has large within-class variation in the image level. One
contributing factor to the within-class variation is cluttered image background. The existing CNN
training uniformly samples image windows from a target image, acting as blind on where is the
foreground (namely, the object of interest) and where is background in the image. While background
portion sometimes benefits base-class image classification by providing context cues, it is less
likely to help fine-grained image classification because all the categories belong to the same base
class (like the car class in this case). Figure~\ref{fig:pull_fig} shows our proposed pipeline for
fine-grained image classification. In contrast to the uniform sampling as used in most existing CNN
training approaches, we first apply detection on a target image to derive the location of the object
of interest and then use the location information to enable an \emph{object-centric sampling} (OCS)
scheme. The OCS tends to sample more image windows around the detected object for CNN training. 

In fact, it is an obvious idea to use object detection to help fine-grained image classification to
down-weight cluttered background, but there has not been much success in the literature in this
direction. To our view, there are two critical requirements for object detection to help: 1)
accurate object detection, and 2) a robust mechanism to handle imperfectness in detection results.
To address the first requirement, we introduce a \emph{saliency-aware object detection}, which
consists of the efforts in both constructing a saliency-oriented dataset and training a
saliency-aware detector. For dataset construction, we only label the most salient object as the
ground truth for an image while both the background and less salient objects (like smaller objects
or occluded objects) in the image are labeled as negative samples. We adopt the Regionlet object
detection framework~\cite{WangICCV13} to learn our object detector because of its unique
characteristics that it operates on original images and has the capability of implying object scales
through detection response, as discussed in Sec.~\ref{sec:regionlet}. Indeed, we exploit the
specific property in the fine-grained image classification setting -- that is, the object of
interest is always the most salient object in an image â€“ to achieve high accuracy in object
detection. To address the second requirement, we propose the OCS to be a multinomial sampling with a
soft emphasis over the detection output. The training samples which have larger overlap with the
detection or closer to the original image would have a higher probability to be sampled during CNN
training. On one hand, the OCS incorporates detections to implicitly down-weight noisy samples from
irrelevant background; on the other hand, it is robust to imperfect detections because of soft
sampling scheme. 
 

The major contributions of this paper lie in three folds: 1) we introduce a large-scale fine-grained
car classification dataset to achieve strong baseline performance using CNN, enabling the study of
the approaches for further improving CNN; 2) based on the Regionlet framework, we propose a
saliency-aware object detection method that is specifically tailored for detection in a fine-grained
image classification setting; 3) we propose an object-centric sampling (OCS) scheme to replace
uniform sampling in CNN training, and the OCS is a multinomial sampling for handling the
imperfectness in detection results. Our proposed overall approach achieves significant performance
improvement, improving the top-1 classification accuracy of 81.6\% by the state-of-the-art CNN to
89.3\% on the large-scale car dataset.

The rest of this paper is organized as following: Sec.~\ref{sec:related_work} describes the related
work. Sec.~\ref{sec:dataset} introduces the large-scale fine-grained  car classification dataset.
Sec.~\ref{sec:ocCNN} presents our saliency-aware object detection method and the OCS scheme for CNN
training. Sec.~\ref{sec:exp} discusses the experiment results, followed with conclusion.

\section{Related Work}
\label{sec:related_work}

\paragraph{\bf Deep convolutional neural network for image classification.} Deep convolutional
neural network (CNN)~\cite{lecun1995convolutional} has become the dominant approach for large-scale
image classification. Since Krizhevsky \emph{et. al.}~\cite{Alex2012} overwhelmingly won the
ImageNet Large Scale Visual Recognition Challenge using a deep CNN, CNN has been widely used for
large-scale image classification tasks. There are efforts to improve CNN architecture, for example,
recent works of using more layers ~\cite{GoogleNet2014}, to achieve even better performance. The
effort of this work is orthogonal to those efforts. Rather, we will use the CNN proposed in
~\cite{Alex2012} as the example, and we show how to incorporate object detection results to improve
classification accuracy. The approach proposed in this paper, object-centric pooling (OSC), is
expected to be applicable to CNNs with deeper architectures~\cite{GoogleNet2014, Simonyan2014}.

\paragraph{\bf Fine-grained image classification datasets.} CNN has been known to work well on
large-scale classification datasets, but it is often suffered in the case of small training data due
to over-fitting. Unfortunately, in the research community, most of the existing fine-grained image
classification benchmark datasets are fairly small. This is because fine-grained labels are often
difficult to obtain. Table~\ref{tab:datasets} lists the sizes of some popular fine-grained image
classification benchmark datasets.  To enable applying CNN for fine-grained image classification, we
construct a large-scale fine-grained car classification dataset. The dataset consists of 333 car
classes with 157,023 training images and 7,840 testing images. With this large-scale dataset, CNN is
able to achieve 81.6\% top-1 classification accuracy, which serves as a strong baseline for us to
further study the possible approaches that could go beyond the current CNN.

\begin{table}
	\caption{The size of our car dataset in comparison with the sizes of existing fine-grained image classification benchmark datasets.
In term of the number of training images, our car dataset is more than one magnitude larger than existing datasets.
}
	\begin{tabular}{ | l | c | r | r |}
		\hline
		\multicolumn{1}{ |c }{\multirow{2}{*}{Datasets} } & \multicolumn{1}{ |c }{\multirow{2}{*}{Classes} } & \multicolumn{2}{ |c| }{\# of Images} \\ \cline{3-4}
		& & Train & Test \\ \cline{1-4}
		Caltech-USCD Bird~\cite{WahCUB2011} & 200 & 5,994 & 5,794 \\
		Oxford Flower 102 ~\cite{NilsbackZ08} & 102 & 2,040 & 6,149 \\
		Stanford Dog~\cite{khosla2011novel} & 120 & 12,000 & 8,580 \\
		Oxford Cat\&Dog~\cite{ParkhiVZJ12} & 37 & 3,680 & 3,669 \\
		Stanford Car~\cite{KrauseICCV13} & 196& 8144& 8041\\
		Our car dataset & 333 & 157,023 & 7,840 \\
		\hline
	\end{tabular}
	\label{tab:datasets}
\end{table}

\paragraph{\bf Fine-grained image classification.} Fine-grained image classification has been an
active research topic in recent years. Compared to base-class image classification, fine-grained
image classification needs to distinguish many similar classes with only subtle differences among
the classes. There has been much work ~\cite{ZhangCVPR12,DengCVPR2013,yang_nips12} aiming at
localizing salient part of fine-grained classes. To ease the challenge, many of them even assume
that the ground-truth bounding boxes of the objects of interest are given. This work is different in
two aspects.  First, rather than using ground-truth bounding boxes, we make attempts to train a good
object detector by proposing a saliency-aware detection approach based on the Regionlet framework.
Second, we build a mechanism to handle imperfect detection results.

\paragraph{\bf Object centric classification.} Object centric classification means using object
location information for image classification. It is different from popular approaches like spatial
pyramid matching (SPM) where an image is blindly divided into SPM blocks for feature pooling.
Apparently, if the accurate information about object location could be obtained, object centric
classification should be a better choice than SPM based classification. This is especially the case
for fine-grained image classification where the key is to distinguish subtle differences from
similar classes and removing cluttered background is helpful. This work is conceptually similar to
object centric pooling (OCP) as in ~\cite{RussakovskyECCV12}, but it is different in the way that we
designed a much more powerful detector to achieve higher detection accuracy. This is done by
exploiting some specific properties that exist in fine-grained image classification. And, the
detector here is trained in a supervised manner. In contrast, in OCP, the detectors were trained in
an unsupervised manner, and the resulting detectors were fairly weak.

\paragraph{\bf Detection with Regionlets.} There is a rich literature in object detection research.
Deformable part model (DPM)~\cite{Felzenszwalb08} has been a popular approach for generic object
detection in the past years. Recently, regions with CNN (R-CNN) approach~\cite{RossCVPR14} achieves
excellent performance on benchmark datasets. Both approaches require to scale images (so that the
object is fit into a fixed-size sliding window) or warp candidate bounding boxes (to the same size
to be input into CNN). Such treatments enable scale-invariant property. However, in the case of
object detection for fine-grained image recognition, scale is an important saliency cue that we hope
to exploit, as explained in more details in Section\ref{sec:regionlet}. Regionlet
approach~\cite{WangICCV13} is a good choice because it operates on the candidate bounding boxes
proposed on original images, and it has the capability to utilize scale as an important saliency
cue. This work also makes some important modifications to the original Regionlet approach, namely,
saliency-aware object detection, which exploits the special property in the setting of fine-grained
image classification where the object of interests is always the most salient (e.g., not occluded,
occupying a big portion of image, etc) object in an image.

\section{A Large Scale Car Dataset}
\label{sec:dataset}

\begin{figure}[h]
	\begin{center}
		\includegraphics[width=1.0\linewidth]{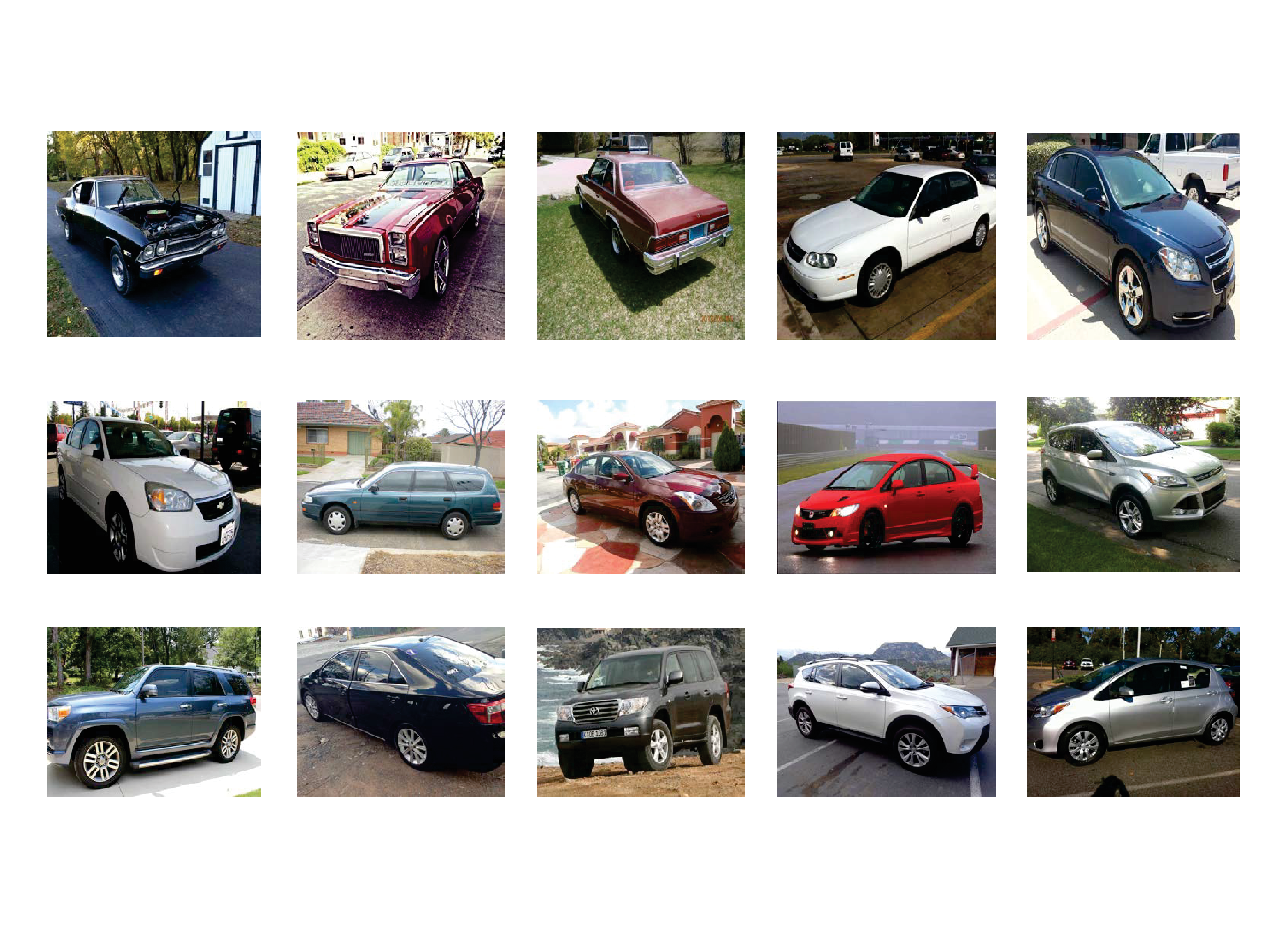}
		\caption{Some sample images from our large-scale fine-grained car classification dataset.
			Each image is labeled as maker, model and year blanket, for example, Chevrolet Equinox
			2010$\sim$2013. The use of year blanket is because the cars within the range of the same
			year blanket do not change their designs on outlook and thus they are visually not
			distinguishable. All images are naturally taken images. The objects of interest are
		mostly centered in images, and they have fairly arbitrary poses.}
	\label{fig:car_sample}
	\end{center}
\end{figure}

To construct a large-scale fine-grained car classification dataset, we crawled images from Internet
and hired car experts to provide a label for each image. We first tried to use Amazon Mechanical
Turk to label the images, but the returned fine-grained labels were too noisy to be useful.
Therefore, we ended up hiring car experts to label the images in-house. We also purposely discarded
artificially synthesized images and only kept the natural images directly taken by cameras. We also
paid more attention to images where a car of interest is the most salient object in an image. After
8 months of efforts of 4 full-time labelers, we have obtained 157,023 training images and 7,840
testing images from totally 333 fine-grained car categories. We made efforts of cross-checking among
different labeler to ensure high-quality labels, and the images in the testing set was ensured to
have no overlapping with the images in the training set.

\begin{figure}[t]
\begin{center}
	\subfloat[Number of cars versus manufacture years.]
	{\includegraphics[width=0.63\linewidth]{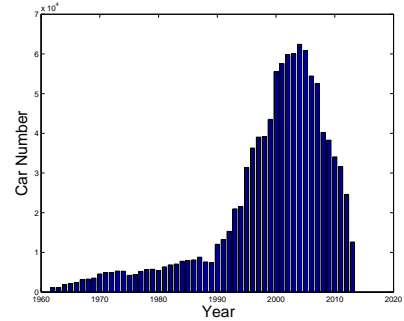} }\\
	\subfloat[Number of images in each of the 333 fine-grained
	categories.]{\includegraphics[width=0.63\linewidth]{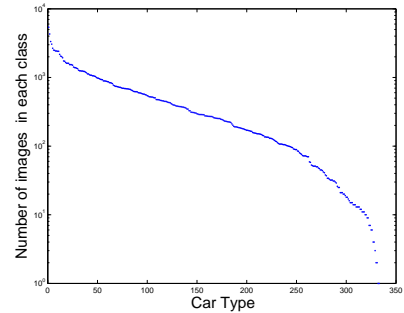} }
	\caption{Some statistics of our fine-grained car dataset.}
\label{fig:cardataset}
   \end{center}
\end{figure}

\begin{figure}[th]
\begin{center}
   \subfloat[Histograms of the relative object size versus image size in the fine-grained car
   dataset]{\includegraphics[width=0.63\linewidth]{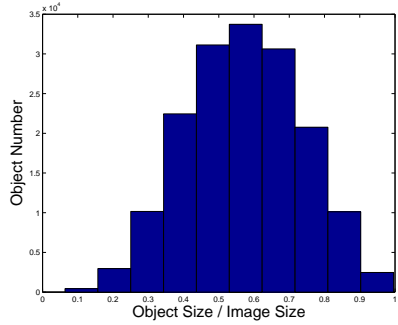} }\\
   \subfloat[Histograms of the relative object size versus image size in the PASCAL VOC 2007
   dataset]{\includegraphics[width=0.63\linewidth]{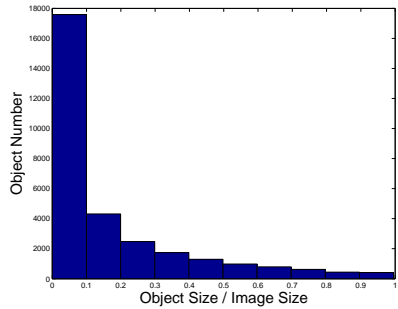} }
   \caption{Statistics of object size versus image size.}
\label{fig:size_diff}
   \end{center}
\end{figure}

The dataset currently covers most car types from 5 brands (or makers), \emph{Chevrolet},
\emph{Ford}, \emph{Honda}, \emph{Nissan} and \emph{Toyota}. Figure~\ref{fig:car_sample} shows some
sample car images from the dataset. There are totally 140 different car models with years ranging
from 1962 to 2013. Each of the car images is labeled to be one (and only one) of the 333
fine-grained
car categories. Table~\ref{tab:car_dataset} describes the divides of the dataset with respect to
car makers. Figure~\ref{fig:cardataset} shows some statistics of the dataset.

\begin{table}[h]
	\begin{center}
		\footnotesize

		\caption{ The statistics of car images versus makers.}
		\label{tab:car_dataset}

		\setlength{\tabcolsep}{4.5pt}
		\begin{tabular}{ l | c|c|c | c}

			\toprule

			{\bf Manufacturer } & {\bf total \#} & {\bf \# of models}& {\bf \# of years} & {\bf \#
			of categories}\\
			\hline
			Chevrolet & 47909 & 25 & 52 & 66\\
			\hline
			Ford& 13542 & 31 & 38& 69\\
			\hline
			Honda& 41264 &24 & 24& 61\\
			\hline
			Nissan& 24611 &37 & 44& 76\\
			\hline
			Toyota& 37537 & 23& 49& 61\\
			\hline

		\end{tabular}
		\label{tab:compare_SOA}
	\end{center}
\end{table}


It is important to realize that the images for fine-grained classification are very different from
the images for generic base-class classification (like PASCAL VOC). This is because, when a user
takes an image for the purpose of fine-grained image classification, the user is more or less in a
collaborative mode to take a close-up photo. This is particularly the case when we are thinking of
the scenario of \emph{search by image} where an user tries to take a photo by a smartphone and then
search for relevant information on Internet (for example, the price of the same type of car on
Craigslist). Figure~\ref{fig:size_diff} shows the histogram of the relative sizes of the objects of interest in
the images from the fine-grained car dataset, and it is contrast by the case of PASCAL VOC 2007. For
the fine-grained car dataset, it is hard to label bounding boxes on all images. As a result, we
sampled about 11,000 images from the dataset and manually labeled bounding boxes on the objects of
interest in those images. From Figure~\ref{fig:size_diff}, it is evident that, for fine-grained image
classification, an object of interest often occupies a significant portion of an image, reflecting
its strong saliency in an image; but it is not the case for base-class classification as in PASCAL
VOC.

While we are using our available dataset for this work, we continue to grow our car dataset, both
covering more car categories and enriching the existing classes that currently have too few training
images.

\section{Object-centric Sampling for Fine-grained Image Classification}
\label{sec:ocCNN}

\subsection{Saliency-aware Object Detection}

The term ``saliency'' is usually referred as the saliency map which describes how salient a pixel is
in the image. Without any confusion, here we borrow the term ``saliency'' to describe the importance
of an object in an image for fine-grained image classification. Thus the ``saliency'' is defined in
object level in contrast to pixel level.

\begin{figure}[t] 
	\begin{center}
	\includegraphics[width=0.85\linewidth]{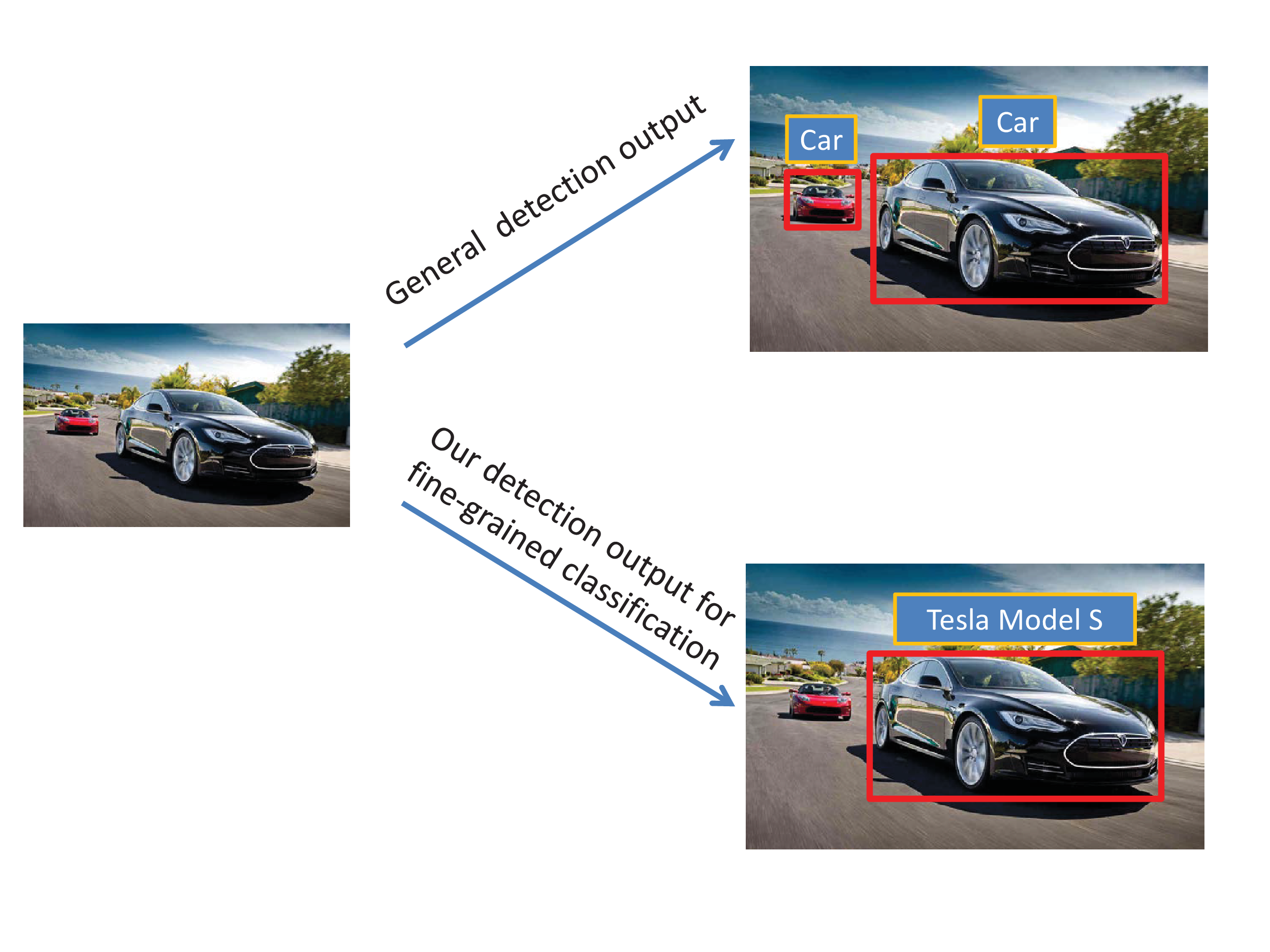} \caption{Difference between object
		detection for fine-grained image classification and general object detection. Our detection
		problem is defined to find the most salient object target in the image, \ie only the most
	salient detection is used if there are multiple detections.} 
\label{fig:det_diff} 
\end{center} 
\end{figure}

The target of object detection for fine-grained image classification is different to that of general
object detection. In later case, we aimed at localizing all the objects of interest. While in
fine-grained image classification, usually there is only one object that represents the fine-grained
label of the image. As shown in Fig~\ref{fig:det_diff}, the most salient object generally
corresponds to the fine-grained label if multiple objects exist. Thus small detections are less
likely to be the required detection compared to bigger detections. If two detections have the same
scale, completely visible objects are more likely to be of interest than significantly occluded
objects. These fundamental differences put specific requirements on the object detector and the
training strategy. The object detector should be aware of object scales and occlusions, \etc.
Ideally, small detection responses should be linked to relatively small or occluded objects, or
false alarms.  We resolve the challenge by constructing a saliency-aware dataset and using a
scale-aware object detector. The occlusion awareness is implicitly achieved by training the detector
with visible objects. 

\subsubsection{Construct training/testing set for detection}

We construct a saliency-aware training set for our object detector. As aforementioned, traditional
detection labeling encourages to detect all objects appearing in an image, which may not comply with
the task of fine-grained image classification. To facilitate saliency-aware detection, we only label
the salient object in one image and surely it should be consistent with the fine-grained category
label, \ie the labeled object should belong to the fine-grained category. For each image, we label
one and only one object as the detection ground truth. When multiple instances are available, the
selection is done based on a mixed criteria of saliency: 

\begin{itemize}
	\item The bigger object is preferred over smaller objects.  
	\item The visible object is preferred over occluded objects. 
	\item An object in the center is preferred over objects around the corner.
	\item The object's fine-grained category label is consistent with the image label.
\end{itemize}
Typically only one object satisfies one or more of these criteria. In any case multiple instances
equally meet these criteria, which is not likely to happen, a random object is selected for the
ground truth labeling.

Note that there might be small cars, occluded cars, cars off the center that are given negative
labels. We delicately apply this labeling protocol to enhance the saliency-aware training.  As a
consequence, the smaller or occluded cars are likely to have relatively smaller detection response
because they have higher chance being put into the negative set. For middle scale objects which
could appear in positive samples for some images and in negative samples for others, we rely on the
object detector to produce a ``middle'' high score. 

Labeling all the images in the large-scale dataset is very expensive and not necessary. We totally
labeled 13,745 images, in which 11,000 images are used for training and 2,745 images are used for
testing. It corresponds to slightly more than 8\% of the entire fine-grained car dataset. 

A detector trained on the constructed dataset is not necessarily capable of detecting the salient
object. For example, the most import saliency factor, scale of the object, is not distinguishable
based on detection responses for most object detectors. Because only one detection can be used for
our fine-grained image classification, we have to struggle with the option between a bigger size
detection with lower response and a smaller size detection with higher detection response if the
detector does not imply the object scale by detection scores.  

\subsubsection{Train the scale-aware detector: Regionlet}
\label{sec:regionlet}

Most of the popular object detectors~\cite{xiaoyu09,Dalal05,Felzenszwalb08,RossCVPR14} are learned
with a fixed scale template.  For example, Dalal \etal~\cite{Dalal05} and Wang \etal~\cite{xiaoyu09}
trained the human detector using a $96 \times 160$ HOG template. The root filter and part filters in
deformable part-based model~\cite{Felzenszwalb08} also use fixed resolution HOG templates. Deep
convolutional neural network based object detectors such as ~\cite{RossCVPR14} are learned mallith a
fixed input resolution such as $224\times 224$ or $227\times227$. Although multiple scale object
detection can be achieved by operating the learned models on different scales of the image, these
detectors can not distinguish object scales. In other words, they give the same response to two
objects only differed in scales, because these two objects will be resized to the same scale and fed
to the detector.  

The Regionlet detector introduced by Wang \etal~\cite{WangICCV13} directly deals with the original
object scale. It supports training and testing on arbitrary detection windows generated from
low-level segmentation cues. In contrast to warping positive samples to a fixed resolution, the
Regionlet approach takes the original positive samples as the input, thus preserving the scale
information during training. The Regionlet classifier is a boosting classifier composed of thousands
of weak classifiers:
\begin{equation}
	H(\x) = \sum_{t=1}^{T}h_t(\x),
\end{equation}
where $T$ is the total number of training stages, $h_t(\x)$ is the weak classifier learned at stage $t$
in training, $\x$ is the input image. The weak classifier $h_t(\x)$ can be written as a function of several parameters:
the spatial location of Regionlets in $h_t$, and the feature used for $h_t$, as following: 
\begin{equation}
	h_t(\x) = G(\p_t, f_t, \x),
\end{equation}
where $\p_t$ is a set of Regionlet locations, $f_t$ is the feature extracted in these regionlets. The
feature extraction locations $\p$ are defined to be proportional to the resolution of the detection
window.
Because feature extraction regions are automatically adapted to accommodate the detection window
size, the Regionlet detector operate on the original object scale. Thus we use the Regionlet
detector for our fine-grained image classification. 

In testing phase, we apply the Regionlet detector to all the object proposals. We extend the
conventional non-max suppression by only taking the object proposal which gives the maximum
detection response. This operation is done over the whole image, regardless of the overlap between
two detections.

\subsection{Object-centric sampling for CNN training}

With  the detected bounding boxes, our next question is how to utilize these bounding boxes to train
an accurate deep CNN for fine-grained classification. Most previous studies~\cite{ZhangCVPR12}
exploiting the bounding box annotations simply crop the image patch  within a bounding box  and use
the cropped image (probably resized) as the input to a learning system. However, this method suffers
from a crucial deficiency: the detected bounding boxes might not be hundred percent correct.
Therefore cropping the image according to the detected bounding boxes may loss a lot of useful
information. On the other hand, translation is  an effective technique for data augmentation for
preventing over-fitting of the deep CNN. Therefore, we do not crop out a single image  but rather
generate multiple patches guided by detection. 

The sampling based approach has been exploited to generate multiple image patches from an image for
data augmentation. The common practice is to generate an image patch\footnote{A fix sized image
patch is full determined by its starting coordinates in the left upper corner.} by random sampling
over a valid range, which is also implemented by the popular
Cuda-convnet2~\footnote{\url{https://code.google.com/p/cuda-convnet/}}.  However, uniform sampling
has innocently ignored the position of the interesting object and  it could end up with many sampled
patches with small overlap with the interesting object, consequentially confusing the learning of
CNN. To be more effective in utilizing the detected bounding boxes, we propose in the paper a
non-uniform sampling approach based on the detected position of the interesting object. The
assumption of the non-uniform sampling is that the detected bounding box provides a good estimation
of the true position of the interesting object. The further of an image patch from the detected
region, the less likely it will contain the interesting object. To this end, we generate multiple
image patches with a given size according to how much they overlap with the detected region. 

\begin{figure}[t]
	\begin{center}
		\subfloat[Object-centric
		sampling]{\includegraphics[width=0.45\linewidth]{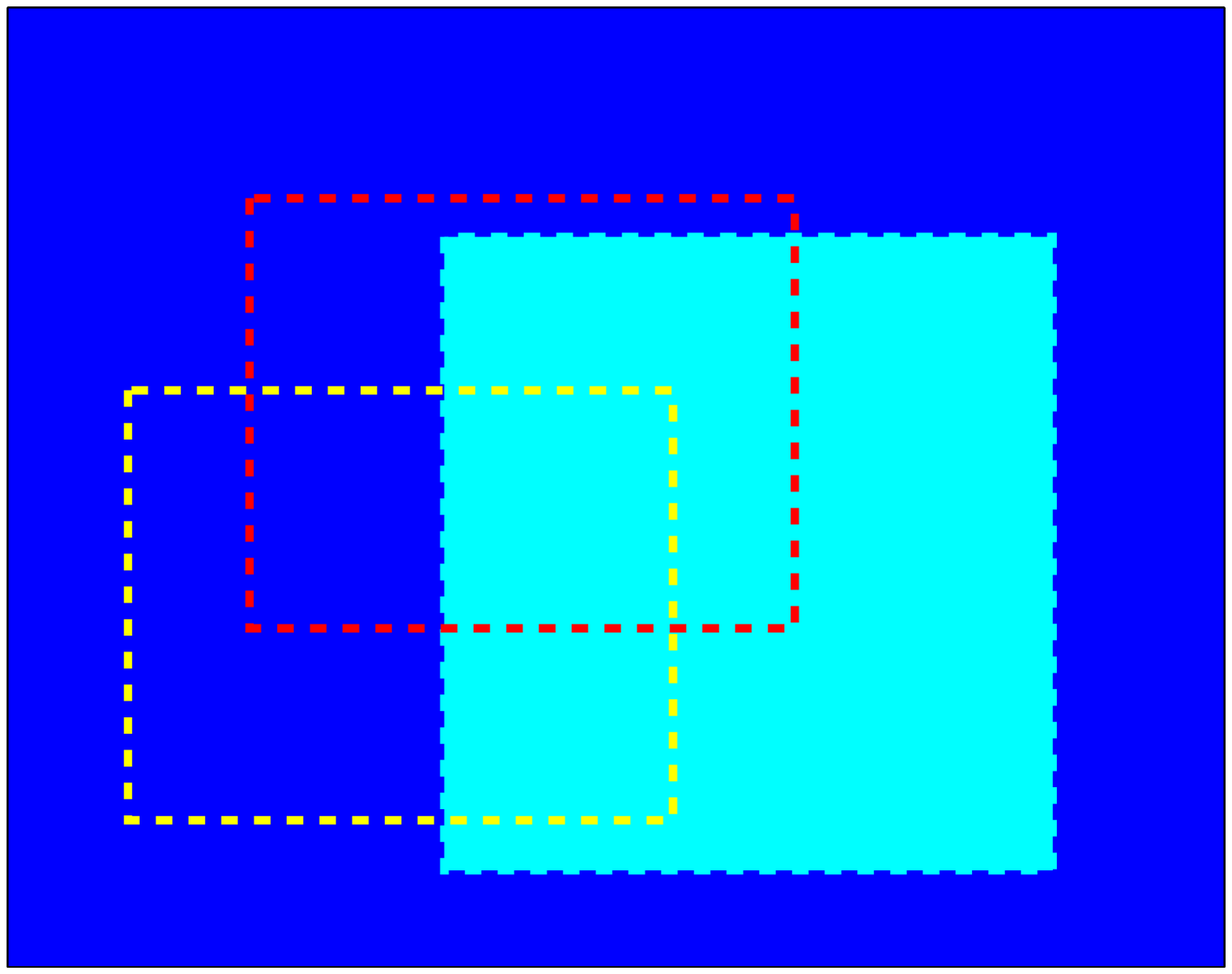}}
		\subfloat[Probability map]{\includegraphics[width=0.45\linewidth]{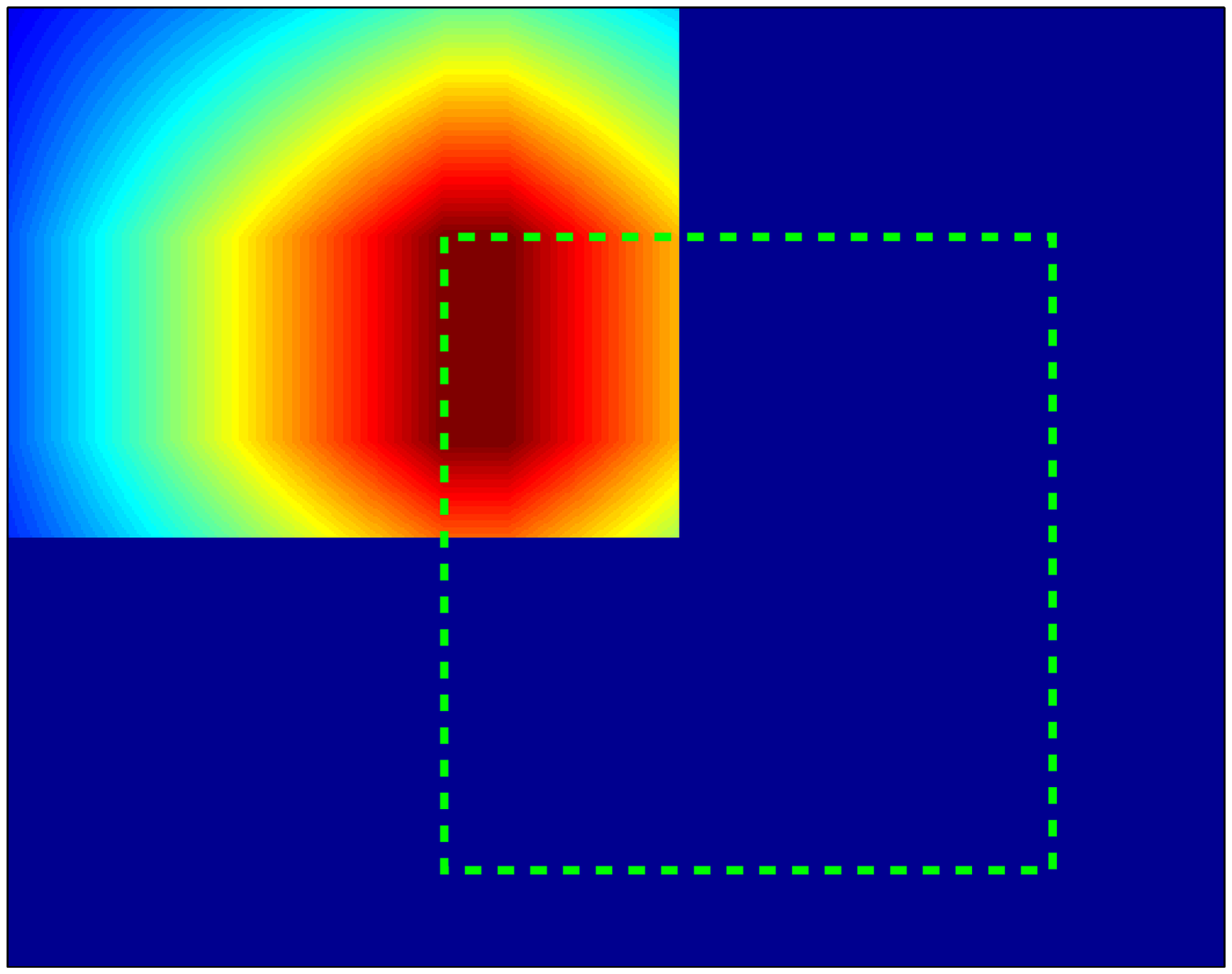}}
				 \end{center}
				    \caption{Left: an illustration of object-centric sampling. The largest rectangle 
						corresponds to the image. The cyan region corresponds to the detected
						bounding box and the red dashed square corresponds to a candidate patch
						sampling. Due
						to that the red square has a large overlap with the cyan region it has a
						large probability to be sampled compared to the yellow square. Right: an
					illustration of the probability map (upper left square). }
					\label{fig:query}
				\end{figure}

Let $s*s$ denote the  size of the input image to CNN, which is also the size of the sampled image
patch.  Given a training image $I$ with size $w*h$,  we let $(x_o,y_o)$ denote the coordinate of the
detected  object, i.e., the center of the bounding box that includes the interesting object and let
$\mathcal R_o$ denote the region of the detected bounding box. Similarly, let $(x, y)$ denote a
position in the image and it is associated with a fixed size $s*s$ region $\mathcal R_{x,y}$ that is
centered at $(x,y)$. The sampling space is given by $\mathcal S=\{(x,y): \mathcal R_{x,y}\subset I,
|\mathcal R_{x,y}\cap \mathcal R_o|\geq \tau \}$, where $\tau$ is an overlapping threshold and
$|\mathcal R_{x,y}\cap \mathcal R_{o}|$ denotes the size of overlap  between the image patch defined
by $(x,y)$ and the bounding box. We set $\tau$ to be 0 and sample $(x,y)\in\mathcal S$ following a
multinomial distribution, with a probability proportional to $|\mathcal R_{x,y}\cap \mathcal R_o|$.
Thus, a region with a large overlap with the bounding box has a high probability to be sampled and
used as a training example to the CNN. This is illustrate in Figure~\ref{fig:query}. 

In order to efficiently implement the multinomial sampling of image patches, we can first compute a
cumulative probability map for each training image according to the detected bounding box and then
sample a coordinate by uniform sampling from the probability quantiles. The prediction on a testing
image are averaged probability over five crops from the original image and their flipped copies, as
well as five crops around the detection and their flipped copies.   

\section{Experiments}
\label{sec:exp}

\subsection{Dataset}

We carried out experiments on our large-scale car dataset. For the evaluation of object detection
performance, we follow the PASCAL VOC evaluation criterion, but increase the requirement of overlap
with ground truth to be 80\%. We view 50\% overlap as too much offset from the ground truth
for fine-grained image classification.

\subsection{Car Detection}

We use selective search~\cite{VanSelectiveSearch} to generate object proposals for detector training
and testing. In the training, object proposals which have more than 70\% overlap with the ground
truth are selected as positive samples. Object proposals which have less than 30\% overlap with the
ground truth are used as negative training samples.  To further polish the localization precision,
we use the Regionlet Re-localization method~\cite{WangACCV14} to learn a support vector regression
model to predict the actual object location.

\begin{figure}[t]
   \includegraphics[width=1.0\linewidth]{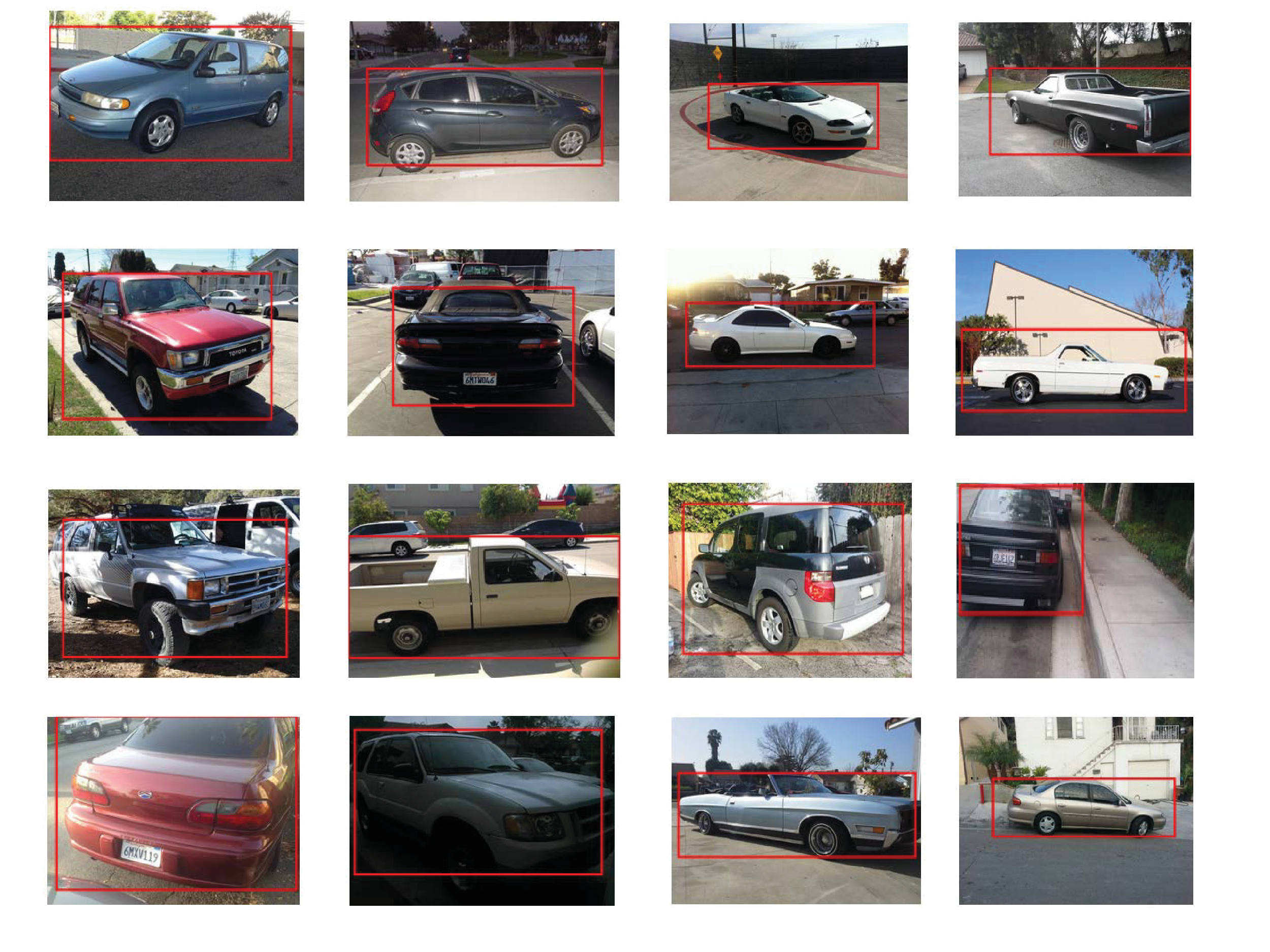} 
   \caption{Example detections on the large-scale car dataset. The detector is quite robust to
   multiple viewpoints. It gives the highest score to the salient car if there are multiple cars in
   the images.}
   \label{fig:det}
\end{figure}

We use 11,000 images to train our car detector. Our training procedure follows ~\cite{WangICCV13}.
The final car detection model has 8 cascades with around 10 thousand weak classifiers in total. As
aforementioned, we replace the non-max suppression post processing scheme with a max-operation over
the whole image.Our detector achieves 85.8\% detection average precision. Our experiment
demonstrates that detection for fine-grained image classification is doable if the training dataset
and the detector are carefully designed. Figure~\ref{fig:det} shows sample detector responses on the
detection testing dataset. 

\begin{figure}[t]
	\centering
   \includegraphics[width=0.9\linewidth]{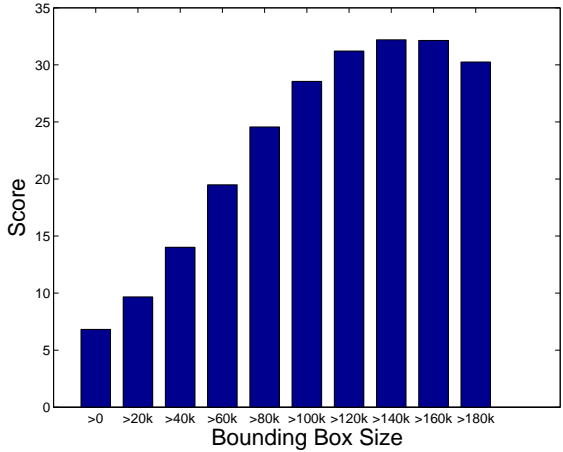} 
   \caption{Average detection scores for objects in different sizes. The horizontal axises shows
   object size in number of pixels. The vertical axises shows average detection scores for objects
   in corresponding sizes. Our detector generally gives larger score to bigger objects.}
   \label{fig:det_score}
\end{figure}

To validate whether the detector is able to produce more confidence detection for relative large
objects, which is crucial for the following image classification process, we plot the average
detection score versus object size in Figure~\ref{fig:det_score}. It shows that the detection
confidence for relatively bigger objects is consistently higher. 

\subsection{Fine-grained image classification}

We directly utilize the neural network structure for image-net classification as in ~\cite{Alex2012}
except that we have 333 object categories. The fine-grained image classification experiment is
carried out using three different configurations:

\begin{itemize}
	\item {\bf Uniform sampling:} the input image is resized to $256\times y$  or $x\times 256$.
		The $224\times224$ training samples are uniformly sampled from the entire image.
	\item {\bf Multinomial sampling:} the input image is resized to $256\times y$  or $x\times256$.
		The $224\times224$ training samples are sampled from the entire image with a preference to
		the location of the maximum detection response.
\end{itemize}

\begin{figure}[h!]
	\centering
   \includegraphics[width=1\linewidth]{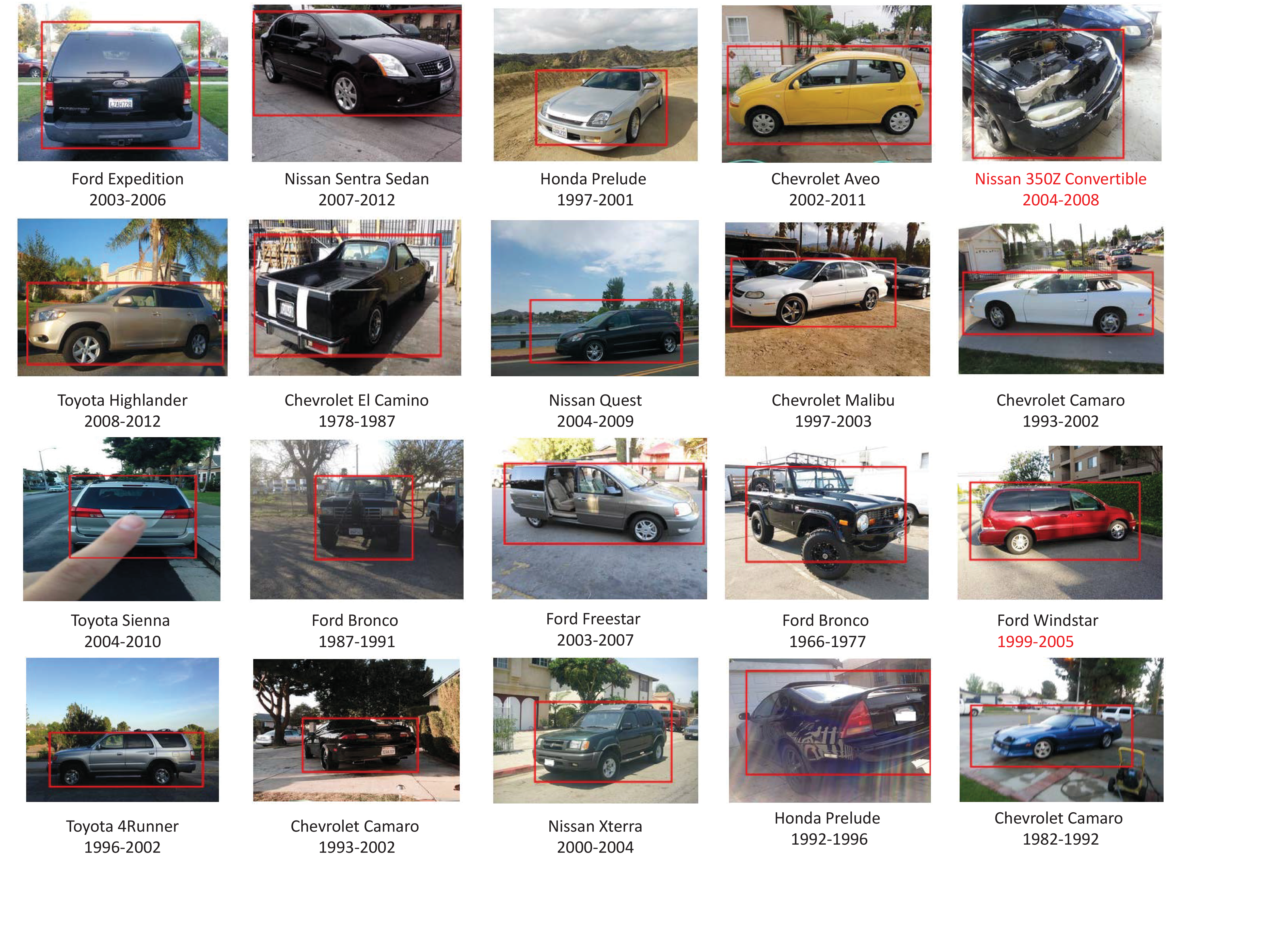} 
   \caption{Sample detection output as well as prediciton results obtained from our object-centric sampling based neural network
	   training. False predictions are colored red. }
\label{fig:rand_crop}
\end{figure}

\begin{table}[h]
	\begin{center}
   \caption{Fine-grained car classification accuracy (\%) with different sampling strategies. Uniform sampling:
	   uniformly sample training images from the entire image. Multinomial sampling: Sample the
	   training images with a probability which is proportional to the normalized overlap between the
	   sample and the detection. }
			\setlength{\tabcolsep}{18.8pt} { \begin{tabular}{ l | c|c}

\toprule

{\bf Sampling method}  & {\bf Top 1 }& {\bf Top 5} \\
\hline
Uniform& 81.6\%& 92.8\%\\
\hline
Multinomial& 89.3\%& 96.6\%\\
\hline
\end{tabular}}
	\label{tab:fine-grained-cls}
\end{center}
\end{table}
The classification accuracy is shown in Table~\ref{tab:fine-grained-cls}. 
The classification accuracy is significantly boosted by enforcing multinomial sampling based on
detection outputs, \ie the top 1 accuracy is improved from 81.6\% to 89.3\%, the top 5 accuracy is
improved from 92.8\% to 96.6\%. Sample prediction results are shown in Figure~\ref{fig:rand_crop}.

\section{Conclusion}

In this paper, we identify the unique properties of fine-grained image classification and delicately
designed an effective pipeline for the challenging task. It includes two techniques: 1)
saliency-aware object detection and multinomial object-centric sampling for deep CNN training. The
first component is achieved by constructing saliency-aware training data construction and training
an adapted Regionlet detector. Compared to traditional detection approaches, our detector yields
higher response on salient objects. The resulting detections are used in an object-centric sampling
scheme to guide the sampling procedure in deep CNN training. The effectiveness of our fine-grained
image classification framework was shown to be dramatic, improving the top-1 classification accuracy
from 81.6\% to 89.3\%. In order to study the effectiveness of the object-centric sampling, we also
constructed a large-scale fine-grained car classification dataset.

Our feature work includes staying the object-centric sampling in CNN with more layers. And we also
continue to build even larger fine-gained car dataset. We are also interested in applying the
proposed framework to other types of objects than fine-grained cars.


{\small
\bibliographystyle{ieee}
\bibliography{hdet,nmetric}
}

\end{document}